\documentclass[journal]{IEEEtran}

\usepackage{graphicx}
\usepackage{amsmath}
\usepackage{amssymb}
\usepackage{multirow}
\usepackage{cases}
\usepackage{algorithm}
\usepackage[noend]{algpseudocode}
\usepackage{graphicx}
\usepackage{amsmath}
\usepackage{amssymb}
\usepackage{color}
\usepackage{soul}
\usepackage{url}
\usepackage[table, dvipsnames]{xcolor}

\def\comment#1{}
\usepackage{enumitem}
\usepackage{wrapfig}

\def\eg{\emph{e.g.}}

\def\ie{\emph{i.e.}}

\def\I{\mathbf{I}}
\def\P{\mathbf{P}}
\def\G{\mathbf{G}}
\def\F{\mathbf{F}}
\def\z{\mathbf{z}}

\def\b{\mathbf{b}}

\begin{document}
\title{Hiding Faces in Plain Sight: Disrupting AI Face Synthesis with Adversarial Perturbations}

\author{Yuezun Li, Xin Yang, Baoyuan Wu and Siwei Lyu,~\IEEEmembership{Senior~Member,~IEEE}%
\thanks{Yuezun Li, Xin Yang and Siwei Lyu  are with the Computer Science Department, University at Albany, State University of New York, NY, USA (e-mail: yli52@albany.edu;xyang8@albany.edu;slyu@albany.edu).}
\thanks{Baoyuan Wu is with the Tencent AI Lab, China (e-mail: wubaoyuan1987 @gmail.com).}
}

\markboth{IEEE TRANSACTIONS ON INFORMATION FORENSICS AND SECURITY}%
{Li \MakeLowercase{\textit{et al.}}: Hiding Faces in Plain Sight: Disrupting AI Face Synthesis with Adversarial Perturbations}

\maketitle

\begin{abstract}
	Recent years have seen fast development in synthesizing realistic human faces using AI technologies. Such fake faces can be weaponized to cause negative personal and social impact. In this work, we develop technologies to defend individuals from becoming victims of recent AI synthesized fake videos by sabotaging would-be training data. This is achieved by disrupting deep neural network (DNN) based face detection method with specially designed imperceptible adversarial perturbations to reduce the quality of the detected faces. We describe attacking schemes under white-box, gray-box and black-box settings, each with decreasing information about the DNN based face detectors. We empirically show the effectiveness of our methods in disrupting state-of-the-art DNN based face detectors on several datasets.
\end{abstract}


\IEEEpeerreviewmaketitle


\section{Introduction}
\IEEEPARstart{T}he recent advances in machine learning and the availability of vast volume of online personal images and videos have drastically improved the synthesis of highly realistic human faces in images~\cite{karras2018style,karras2017progressive} and videos~\cite{Thies_2016_CVPR,suwajanakorn2017synthesizing,kim2018deep,chan2018everybody}, and tools that help to make them\footnote{The most notable example of AI face synthesis system is the freely available FakeApp \url{https://github.com/deepfakes}.}. While there are interesting and creative applications of the AI face synthesis systems, they can also be weaponized due to the strong association of faces to identify an individual. The potential threats range from revenge pornographic videos of a victim whose face is synthesized and spliced in, to realistic videos of state leaders seeming to make inflammatory comments they never actually made or a high-level executive commenting about her company's performance to influence the global stock market.  

\begin{figure}[t]
	\centering
	\includegraphics[width=0.9\linewidth]{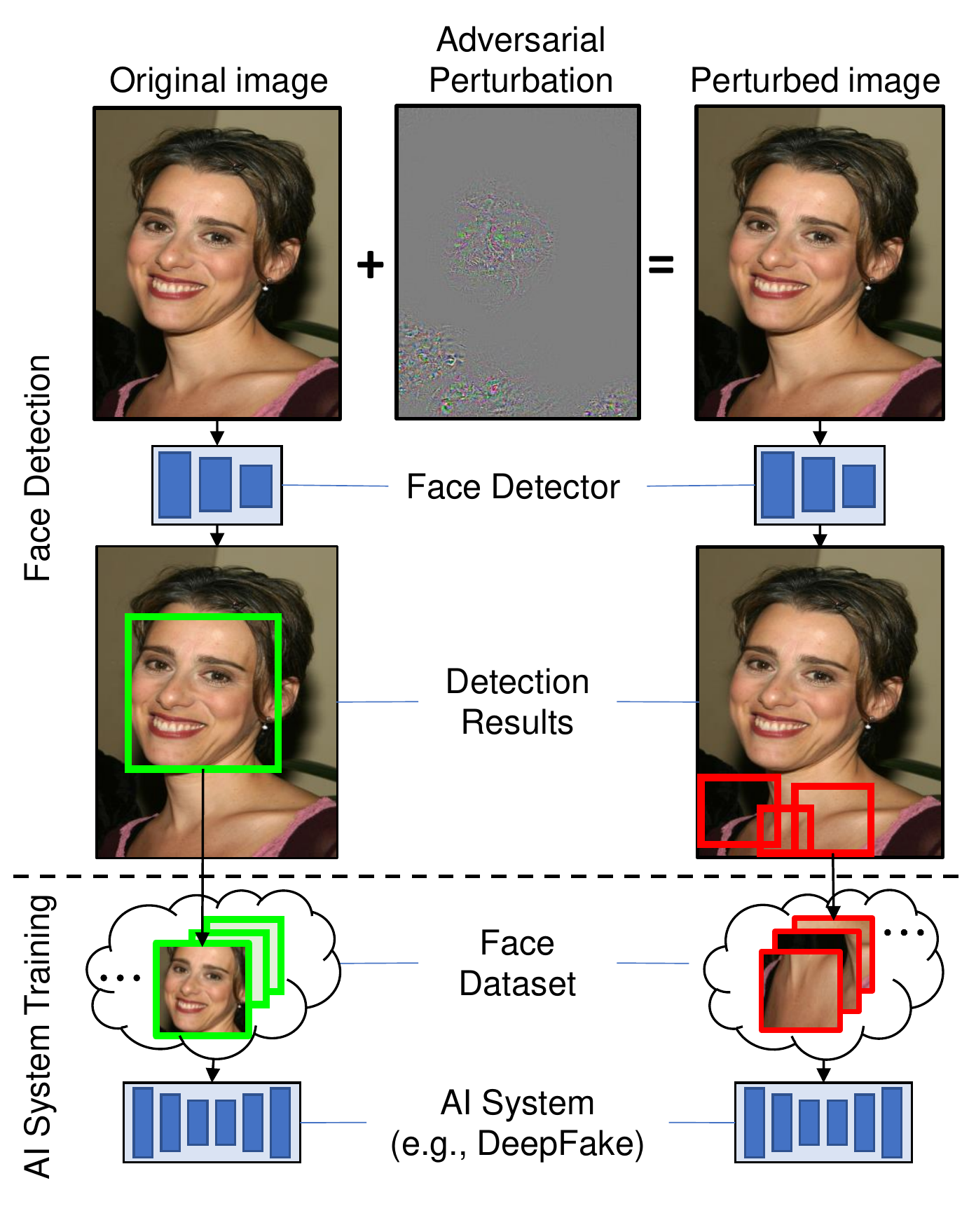}
	\vspace{-0.5cm}
	\caption{\em \small Overview of disrupting AI face synthesis. We aim here is to use adversarial perturbations (amplified by $30$ for better visualization) to distract DNN based face detectors, such that the quality of the obtained face set as training data to the AI face synthesis is reduced.}
	\label{fig:overview}
	~\vspace{-2em}
\end{figure}

Foreseeing this threat, several forensic techniques aiming to detect AI synthesized faces in images or videos have been proposed recently \cite{afchar2018mesonet,li2018ictu,yang2018exposing,guera2018deepfake,li2019exposing,8638330}. However, given the speed and reach of the propagation of online media, even the currently best forensic techniques will largely operate in a postmortem fashion, applicable only after an AI synthesized fake face image or video emerges. On the other hand, current AI face synthesis methods predicate on the availability of {\em face sets} that are automatically detected and cropped faces from an individual's online personal images or videos as training data, Figure \ref{fig:overview}. For effective synthesis, the size of the face set should be sufficiently large, typically in the range of thousands of high resolution faces, with diverse orientations, expressions and lighting conditions. 

As such, a method that can sabotage automatic face detection with fewer actual faces and more non-faces in the resulting face set can significantly slow down the production of AI synthesized faces. This is because if the quality of the automatically detected face set is bad, then faces in the images or videos may have to be manually located. Compared to the passive forensic techniques, this is a more effective {\em proactive} approach to protect individuals from becoming the victims of such attacks. In this work, we study adversarial perturbations to deep neural network (DNN) based face detectors as a countermeasure to AI synthesized faces, which are specially designed signals added to images that are imperceptible to human eyes but can result in detection failures, Figure \ref{fig:overview}. 

We target the DNN based face detectors as they achieve the best face detection performance to date \cite{wang2017face,jiang2017face,sun2018face,yang2017face,najibi2017ssh,zhang2017s3fd,tang2018pyramidbox} as well as improved robustness to variations in pose, expression and occlusion.  They are expect to replace non-DNN based methods in the coming years and become mainstream methods. We start with a new white-box adversarial perturbation generation method, where we have knowledge of internal details of the DNN model including its structure and parameters. The white-box adversarial perturbation generation is cast as a constrained optimization problem, the objective function of which aims to increase the two type of errors for a face detector, namely {\em mis-detections} and {\em false detections}. For a specific face detector, mis-detections correspond to true faces that are not found and false detections are non-faces that are identified as faces. We employ gradient based method to optimize this objective function using back-propagation of the DNN model. 

The applicability of the white-box adversarial perturbation generation scheme may be limited due to the requirements on knowing the network parameters, so we extend to the more general settings where we assume less knowledge about the underlying DNN structure and parameter (\ie, gray-box and black-box settings).  First we describe a randomized gradient update to reduce the dependency on network parameters in generating adversarial perturbations. Furthermore, due to the complexity of the DNN architecture search, existing DNN-face detectors rely on a small number of architectures that have been proven effective, and different face detectors are usually obtained by fine tuning these standard network architectures, we can create adversarial perturbations generated for commonly used base network models for a compound attack taking advantage of the few existing variants of base models. To the best of our knowledge, our method is the first adversarial perturbation generation method dedicated to DNN based face detectors under the gray-box and black-box settings.

Experimental results on three widely used benchmark data sets, \ie, WIDER \cite{yang2016wider}, 300-W \cite{sagonas2013300} and UMDFaces \cite{bansal2016umdfaces} show the effectiveness of our method to reduce the data utility quality in comparison with various state-of-the-art face detectors, and robustness with regards to JPEG compression, additive noise and blurring. Adversarial perturbations targeting face detectors could be used as a user or platform-initiated approach to ``pollute'' the face set of individuals, thus helps to prevent bulk re-use of online personal images/videos as training data to create a synthetic face.

The rest of the paper is organized as follows. {Section \ref{sec:background} covers the background and related works on AI based face synthesis, detection of AI synthesized faces, face detection, adversarial perturbation to DNN models. Section \ref{sec:method} describes the details of our method generating adversarial noise to perturb DNN based face detectors. Section \ref{sec:experiments} describes the experiments we conduct for efficacy demonstration. Section \ref{sec:conclusion} concludes the paper with discussion and future works.}

\section{Background and related works}
\label{sec:background}

\subsection{AI Face Synthesis}

Synthesizing realistic faces using algorithms has been an important task for computer vision and graphics for over three decades. There exists sophisticated computer graphics systems (\eg, 3D Studio Max and Maya) and high resolution 3D surface models of faces to render high-quality realistic human faces. However, the process is lengthy, costly and technically demanding for ordinary users. 

This has been significantly changed with the recent advent of data-driven face synthesis methods based on deep neural networks, which lead to more realistic synthesized faces yet with considerable reduction in time and cost. For example, the {\tt DeepFake} systems generate face replacement videos where one person's face is replaced with synthesized face of the other with the same facial expressions. Whole face or upper-body reenactment is achieved with algorithms such as Face2Face \cite{Thies_2016_CVPR} or DeepPortrait \cite{kim2018deep}. Even more impressive are faces generated completely using generative adversarial networks (GANs) \cite{goodfellow2014generative}, which have the capacity of generating highly realistic facial details \cite{karras2017progressive} and different facial styles \cite{karras2018style}. 

Underneath all data-driven face synthesis methods are machine learning models that need to be trained using large number of detected, cropped and aligned faces that are automatically extracted from online images and videos using state-of-the-art face detection methods.

\subsection{Forensic Detection of AI Synthesized Faces}

Several methods have been recently proposed to detect AI synthesized faces. The method in \cite{li2018ictu} is based on the observation that synthesized face in videos lack eye blinking due to an intrinsic bias of the training face set obtained from image search engine. Another method \cite{yang2018exposing} detects synthesized faces using inconsistent 3D orientations of synthesized faces. DNN based detection algorithms identifies synthesized faces using artifacts introduced in face warping \cite{li2019exposing}, or in the synthesis of eye, mouth or face contours \cite{8638330}, or directly via a trained convolutional neural network  \cite{afchar2018mesonet}. The method of \cite{guera2018deepfake} incorporates a recursive neural network to detect temporal consistency of synthesized faces across different video frames. The work in \cite{rossler2018faceforensics,roessler2019faceforensics++} introduced a large face manipulation dataset in order to alleviate the shortage of training sampling for detecting AI synthesized faces.

Despite these aforementioned forensic methods have achieved various levels of success in detecting images/videos containing AI synthesized faces, they are passive defensive methods that are only applicable after the fake media have propagated online. Given the fast speed of fake images/videos propagate online, these passive defensive methods are not sufficient to combat AI synthesized fake faces. 

\subsection{Face Detection}

Early face detectors examine local image regions in a sliding window fashion for patterns resembling faces. The first efficient and effective face detector \cite{viola2001rapid} uses Haar-type features in a cascaded classifier based on AdaBoost. Subsequently, more robust, effective and efficient face detectors are proposed in the literature based on various feature types such as LBP \cite{ojala2002multiresolution}, SURF \cite{li2013learning, li2011face} and DPMs \cite{ramanan2012face}. Using the HOG feature \cite{dalal2005histograms}, software package {\tt DLib} represents the state-of-the-art for the pre-DNN face detection methods.  

Recently, DNN based face detectors start to becoming mainstream with their high performance and improved robustness with regards to variations in pose, expression and occlusion. Due to the prohibitive cost of optimizing network structures, existing DNN based face detectors concentrate on three common architectures from general object detectors, \ie, RCNN \cite{rcnn}, Faster-RCNN \cite{faster-rcnn} and SSD \cite{liu2016ssd}. Detectors based on RCNN, \eg, \cite{li2015convolutional,farfade2015multi,ranjan2015deep,yang2015facial,yang2015convolutional,ranjan2019hyperface,ranjan2017all}, first identify proposed regions from selective search \cite{uijlings2013selective} and then classify each region proposal as a face or non-face using a RCNN. Faster-RCNN based face detectors \cite{wang2017face,jiang2017face,sun2018face} use Region Proposal Networks (RPN) to generate initial region proposals and is more efficient than those based on RCNN. RPN generates a set of anchor boxes on the image that are potentially faces and predicts the confidence score of being a proposal and bounding box offset for each anchor box in a single forward pass. Then the generated proposals will be further refined to the final face score. Running efficiency of face detection can be further improved with the SSD-based methods \cite{yang2017face,najibi2017ssh,zhang2017s3fd,tang2018pyramidbox}, which locate a set of anchor boxes in the input image, and then predicts the confidence score and bounding box offset. Due to the great performance of VGG \cite{simonyan2014very} and ResNet \cite{he2016deep} architecture, almost all of these face detectors use them as base network.

\subsection{Adversarial Perturbation to Deep Neural Networks}

Adversarial perturbation is a specially constructed noise that induces minimal visual distortion but can change the result of a DNN based algorithm on the perturbed image, taking advantage of the sensitivity of DNNs to the input. 

Many adversarial perturbation schemes have been developed to attack state-of-the-art DNN based image classifiers. In a white-box attack \cite{szegedy2013intriguing,goodfellow2014explaining,kurakin2016adversarial,papernot2016limitations,moosavi2016deepfool,moosavi2017universal,zeng2017adversarial,luo2018towards,baluja2018learning}, the attacker knows the details of the DNN model, including the network architecture and all parameters such as connection weights and biases. Adversarial perturbations can then be generated by minimizing a loss function with respect to the input image aiming to changing the classification results with minimal distortions. Since the details of the DNN model in question are accessible, the gradient of loss function can be computed with a back propagation type algorithm and used to maximize the loss function. In contrast, in a black-box attack \cite{papernot2016transferability,papernot2017practical,liu2016delving,brendel2017decision,ilyas2018black,ilyas2017query}, the attacker has no knowledge of the DNN model. The only information is the final output of DNN model, \ie, the predicted labels or the corresponding confidence score. 

Recently, adversarial perturbation is extended from attacking image classifiers to object detectors, \eg, \cite{lu2017adversarial,xie2017adversarial,eykholt2018physical,chen2018robust,li2018rap}, and specifically, face detectors \cite{bose2018adversarial}. In contrast to image classifiers where only outputs one class label for an image, object detectors have multiple outputs, including the bounding box locations, class label and detection confidence scores. As such, generating adversarial perturbation for object detectors is more complicated than for image classifiers. We cannot directly adopt adversarial perturbation methods designed for image classification to the case of object or face detectors. In particular, we need to extend the definition of gray-box and black-box setting of adversarial perturbation generation for face detection, as described subsequently.

\section{Method}
\label{sec:method}

We describe our method in detail in this section. We start with the method for white-box attack, where we assume access to the DNN model of the face detector. We then generalize the white-box attack to the more general case where the DNN model is not completely visible.   

\subsection{White-box adversarial Perturbation Generation}
\label{sec:white}

\subsubsection{Notations \& Definitions}. We use $[n]$ as a shorthand for set $\{1,\cdots,n\}$. $\I_0 \in [255]^{S_x \times S_y \times 3}$ is a $8$-bit RGB image that contains faces. We use a quadruple $\b = (x,y,w,h)$ to represent a bounding box, where $x,y$ are the left and the top coordinates, and $w,h$ are the width and height of the bounding box. For two bounding boxes $\b = (x,y,w,h)$ and $\b' = (x', y', w', h')$, we compute their {\em intersection over union} (IoU) score as
\begin{eqnarray*}
\textrm{IoU}(\b,\b') & = & \frac{\Delta w \Delta h}{w'h' + wh - \Delta w \Delta h},    \\
\Delta w & = & \min\{x' + w', x + w) - \max(x', x), \\
\Delta h &= & \min(y' + h', y + h) - \max(y', y). 
\end{eqnarray*}
IoU takes values in the range of $[0,1]$, and it is one when $\b$ and $\b'$ completely overlap and zero when they have no overlapping at all. With a slight abuse of notation, we use $\I_0({\b})$ to denote the sub-image of $\I_0$ restricted to bounding box $\b$.

It is a common step in efficient DNN face detectors to first find {\em face proposals}, which are potential candidate regions corresponding to faces. Using face proposals avoids the more expensive sliding window search. We denote $\P  = \{(\b_j)\}_{j=1}^n$ \footnote{In practice, non-maximum suppression (NMS) is usually used to remove nearby proposals corresponding to the same face. Here, we keep all proposals instead, so the adversarial perturbation will affect the final detections.} as the set of bounding boxes of $n$ face proposals obtained on image $\I_0$. For different DNN based face detectors, face proposals are in different forms. The simplest cases are all rectangular regions of fixed sizes in $\I_0$ (\eg, \cite{li2015convolutional,farfade2015multi}). In faster-RCNN based face detector (\eg, \cite{wang2017face,jiang2017face,sun2018face}), the face proposals are generated by region proposal network. SSD based face detectors (\eg, \cite{yang2017face,najibi2017ssh,zhang2017s3fd,tang2018pyramidbox}) does not explicitly generate face proposals but directly adjusts the predefined anchor boxes to detections in a single pass. 

For each face proposal $\b^p_j$, the {\em confidence score} $c_j = \F(\I_0(\b^p_j))$ is computed with the predictor $\F$, which is usually formed in a DNN based face detector using base networks chosen from several widely used architectures, such as VGG \cite{simonyan2014very} or ResNet \cite{he2016deep}. When $c_j$ is greater than a pre-set threshold $\theta_d$ (we choose $\theta_d = 0.5$), in which case $\b^p_j$ is added to the set of detected faces. We define $\G = \{\b^g_i\}_{i = 1}^m$ as the set of the bounding boxes of all detected faces that are treated as {\em ground truth} detections. A face proposal with bounding box $\b^p_j$ is a {\em potential true detection} if $\max_{i \in [m]} \textrm{IoU}(\b^g_i,\b^p_j) \ge \theta_p$ (we choose $\theta_p = 0.3$), and is a {\em potential false detection} otherwise. We denote ${\cal I}^p_\text{td}=\{j|\max_{i \in [m]} \textrm{IoU}(\b^g_i,\b^p_j) \ge \theta_p\}$ as the index set of all potential true detections, and ${\cal I}^p_\text{fd} = [n]\setminus{\cal I}^p_\text{td}$ as the set of all potential false detections out of all face proposals. 

\medskip

\subsubsection{Formulation} The white-box adversarial perturbation generation for image $\I_0$ works with the face proposals $\P$. This is cast as a constrained optimization problem maximizing the cross-entropy loss with regards to the adversarial perturbation image $\z$. Specifically, we solve for
\begin{eqnarray}
\max_{\z} & \!\!\!\!\sum_{j \in {\cal I}^p_\text{td}}  \log (1 - \tilde{c}_j) +  \sum_{j \in {\cal I}^p_\text{fd}}  \delta_{\{\F(\I_0(b^p_j)) \ge \rho\}} \cdot \log \tilde{c}_j    \nonumber  \\
     \textrm{~s.t.~} & \!\!\!\!\! \tilde{c}_j = \F((\I_0+\z)(\b^p_j)), ~~\|\z\|_2 \leq \epsilon. 
\label{equ:total}
\end{eqnarray}
Here $\tilde{c}_j$ is the confidence score of the same face proposal in the adversarially perturbed image $\I = \I_0+\z$ and $\delta_{\{c\}}$ is the indicator function that is 1 when the condition $c$ is true and 0 otherwise. In practice, the number of potential false detections may be large, so we introduce a threshold $\rho$ (we choose $\rho=1000$) to only include those with top confidence scores in the objective function.  The end result of optimizing \eqref{equ:total} is fewer true detections and more false detections will be included in the final detection set. As such, our method effectively reduces the data utility quality of the detected face set when used as training data for AI face synthesis systems. 

\medskip

\subsubsection{Optimization} Solution to Eq.\eqref{equ:total} can be obtained using a projected gradient ascent algorithm. Starting with $\I_0$, at the $t$-th iteration, we update the current estimation of $\I_t$ by first moving it along the direction of the gradient (or sub-gradient when the network involves non-differentiable activaction functions such as ReLU or leaky ReLU) of the objective function of \eqref{equ:total}, which is shorthanded as $L(\z)$, with a small step size $\gamma_t >0$, as 
\begin{equation}
\z_{t+1} = \z_t + \gamma_t \nabla L(\z_t).
\label{e:1}    
\end{equation}
The gradient is computed using the chain rule as  
\begin{flalign}
\!\!\!\!\nabla L(\z) = \sum_{j \in {\cal I}^p_\text{fd}} \frac{\delta_{\{\F(\I_0(b^p_j)) > \rho\}}}{\tilde{c}_j}\frac{\partial F}{\partial \z} - \sum_{j \in {\cal I}^p_\text{td}}  \frac{1}{1 - \tilde{c}_j}\frac{\partial F}{\partial \z},
\label{e:2}
\end{flalign}
where $\frac{\partial F}{\partial \z}$ is the gradient of the DNN model $\F$ with regards to its input computed with the back-propagation algorithm based on the chain-rule \cite{rumelhart1988learning}. Using model gradient makes the search for adversarial perturbation efficient but it also entails a dependence on the details of the underlying DNN model.

Step size $\gamma_t$ is determined to ensure that the update satisfies the constraint as 
\begin{equation}
\gamma_{t} = \arg\!\max_\gamma
\left\{\gamma\left|
\begin{array}{l}
  \|\z_t + \gamma \nabla L(\z_t)\|_2 \le \epsilon \; \& \\
     L(\z_t + \gamma \nabla L(\z_t)) > L(\z_t).
\end{array}\right.\right\}.
\label{e:3}
\end{equation} 
The solution is obtained by a 1D line search procedure. We repeat Eqs.\eqref{e:1}-\eqref{e:3} until the algorithm converges to a solution $\z$, then we generate the perturbed image by projecting $\I = \I_0 + \z$ to the proper RGB value range.  

However, face proposals extracted from the perturbed image $\I$ may be different with those extracted from the original image $\I_0$, so running the aforementioned algorithm directly we cannot precisely affect the final face detection result. To handle this issue, we use a technique known as {\em warm start} \cite{kurakin2016adversarial}. Specifically, instead of running the Eqs.\eqref{e:1}-\eqref{e:3} until convergence, we run one round of the update and obtain a perturbed image $\tilde{\I}$. Then we set $\I_0 = \tilde{\I}$, and run the DNN based face detector $\F$ on the updated $\I_0$ to initiate a new run of  optimization of Eqs.\eqref{e:1}-\eqref{e:3}. This procedure is repeated until either (1) a total number of iterations $T$ has been achieved or (2) no potential true detections in the perturbed images can be extracted (\ie, all true faces have been concealed from the image). The effectiveness of the white-box attack to state-of-the-art DNN based face detectors is experimentally demonstrated in Section \ref{sec:experiments}.

\subsection{Gray-box Adversarial Perturbation Generation}
\label{sec:gray}

To be able to compute the the gradient, Eq.\eqref{e:3}, the white-box adversarial perturbation generation method requires full knowledge of the base network in the DNN based face detector. This limits the applicability of white-box attack, especially when we have no access to the full base network model of the face detector. 

Nevertheless, current state-of-the-art DNN based face detectors rely on two variants of the base network{\footnote{We surveyed $29$ state-of-the-art face detection methods on the leader board of face detection challenge on the WIDER dataset \cite{yang2016wider}, $27$ are DNN based face detectors, and the top performance is achieved from the subset of $23$ methods, using either Faster-RCNN ($5$) or SSD ($18$) as their architecture with VGG or ResNet as base networks.}}, namely VGG \cite{simonyan2014very} or ResNet \cite{he2016deep}, thus the vulnerability of one DNN based face detector is expected to be shared by other DNN based face detectors using the same base network but with different parameters. Therefore, it is possible to extend the white-box adversarial perturbation generation to {\em gray-box} adversarial perturbation generation, where we assume the targeted DNN based face detector has the same base network as a known DNN based face detector. We develop adversarial perturbation generation method based on the latter and apply it to the attack of the former. 

Our solution is to reduce reliance of the adversarial perturbation generation to the exact values of network parameters of a given base network by randomizing the gradient, as 
\begin{equation}
\tilde{\I}_t = \I_t + \gamma_t (\nabla L(\I_t)+{\mathbf{n}_t}),
\label{e:2a}    
\end{equation}
where $\mathbf{n}_t$ is a sample from i.i.d. zero-mean white Gaussian noise with a small standard deviation $\sigma$. This is equivalent to a stochastic perturbation to the gradient, and as the update in Eq.\eqref{e:2a} does not follow the exact gradient obtained from a fixed set of parameters, this measure can extend the white-box attack to a gray-box setting. 

\subsection{Black-box Adversarial Perturbation Generation}
\label{sec:black-box}

The more challenging setting is when we have no knowledge about the base network structure or parameter, other than that the face detector is based on a DNN model. In this work, we define black-box attack to DNN based face detectors as where we have no knowledge of the network structure or the parameters other than that it is based on a DNN model\footnote{Black-box adversarial perturbation generation for classification systems is usually defined for the case where we only have the binary output of the classifier on an input. For face detectors, however, there is not a clear correspondence to that, as face detector outputs not a single label, but a set of bounding box locations.}. 

As mentioned previously, existing DNN based face detectors use two basic types of base networks, and future generation of DNN based face detectors are expected to follow the same trend. So, combining gray-box adversarial perturbations obtained from different DNN based face detectors with known base networks is likely to be effective for the black box attack. Specifically, denote $\F_k$, $k = 1,\cdots, K$, as DNN based face detectors with different base networks, and $L_k(\z)$ as the objective function for individual DNN based face detectors as defined in Eq.\eqref{equ:total}, we find adversarial perturbation for image $\I_0$ by solving
\begin{equation}
\begin{array}{l}
\max_\z \sum_{k=1}^K L_k(\z), ~~\text{s.t.}~~\|\z\|_2 \le \epsilon.
\end{array}
    \label{e:4}
\end{equation}
Eq.\eqref{e:4} is then optimized following the algorithms described in Section \ref{sec:white} and \ref{sec:gray}, with warm start and perturbed gradient update. In subsequent experiments (Section \ref{sec:exp-gray-box} and \ref{sec:exp-black-box}), we demonstrate the effectiveness of gray-box and black-box attacks to DNN based face detectors.

\section{Experiments}
\label{sec:experiments}

\subsection{Experimental Settings}
\label{sec:settings}

\subsubsection{Datasets} 
We validate our method on several widely used data sets. We construct a dataset of $909$ images, which will be referred to as {\em sub-WIDER} subsequently, from the validation set of the WIDER dataset \cite{yang2016wider}, one of the largest benchmark for face detection. In forming this dataset, we exclude faces with small sizes, heavy occluded or unusual orientations as they are not relevant for training AI face synthesis methods.

\begin{figure}[t]
	\centering
	\includegraphics[width=\linewidth]{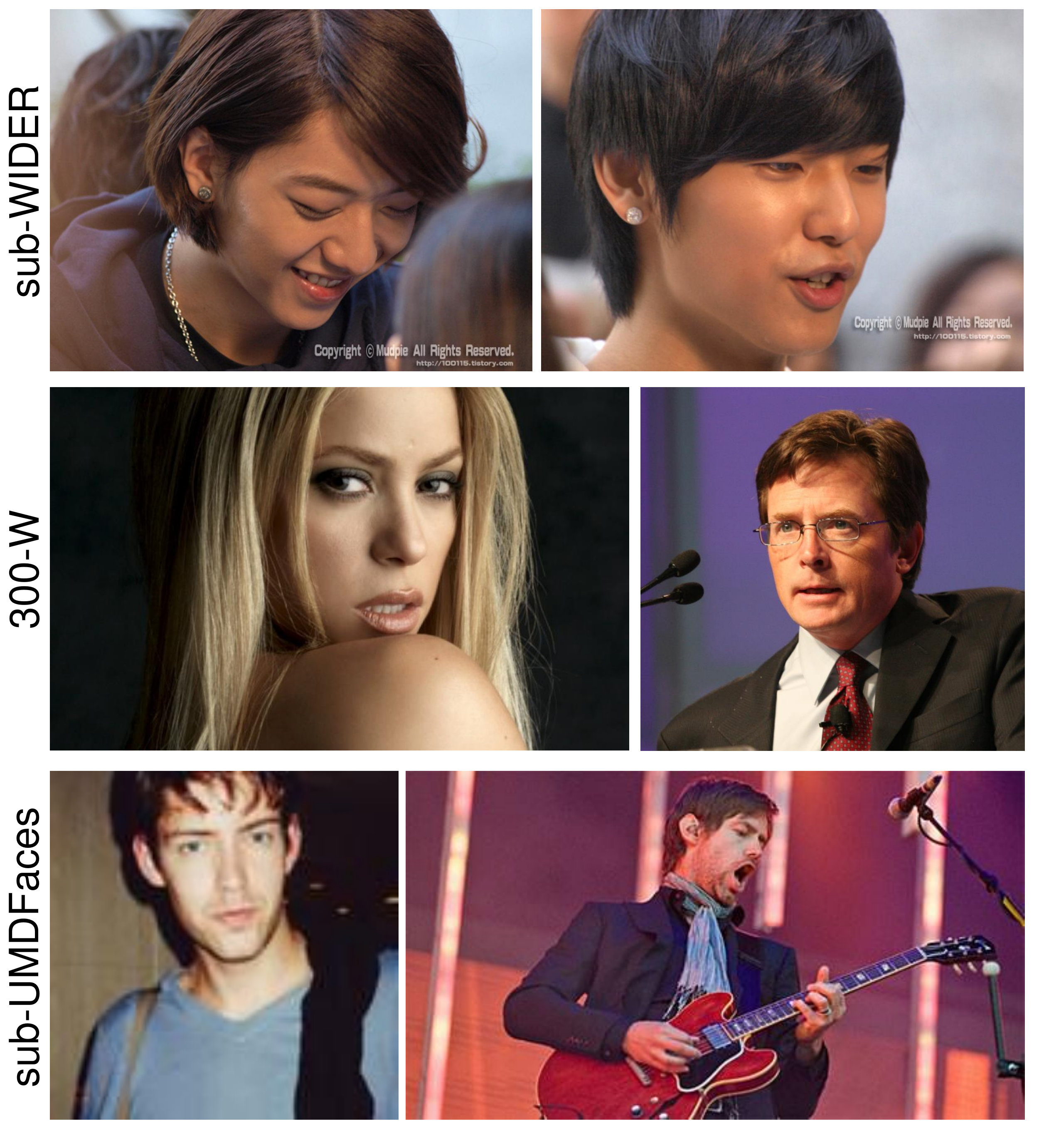}
	\vspace{-0.8cm}
	\caption{\em \small Examples from datasets used in this work.}
	\label{fig:data_example}
	~\vspace{-2em}
\end{figure}

To compare with results of previous works, we also use two other datasets. The {\em 300-W} dataset has $600$ images each containing a single face from the test set of the {\em 300 Faces In-the-Wild Challenge} 300-W \cite{sagonas2013300}\footnote{Since ground truth faces are not labeled in 300-W, we use the detection results of {\tt Dlib} as the ground truth detection, which is also the protocol used in a compared work \cite{bose2018adversarial}.}.  The {\em sub-UMDFaces} dataset is constructed from $500$ images randomly sampled from the UMDFaces dataset \cite{bansal2016umdfaces}. Several image examples are illustrated in Figure \ref{fig:data_example}.
 
\subsubsection{DNN based face detectors}
\label{subsec:face-detectors}
We consider several state-of-the-art DNN based face detectors as the target of our experiments. For Faster-RCNN based face detectors, we consider two different base networks: vgg16 \cite{simonyan2014very} and ResNet101 \cite{he2016deep}, which are denoted by {\em Fv16} and {\em Fr101} respectively\footnote{The Faster-RCNN face detector is based on \cite{jjfaster2rcnn}.}. For SSD based face detectors, we consider two state-of-the-art face detectors: PyramidBox \cite{tang2018pyramidbox} and SFD \cite{zhang2017s3fd}. We use ResNet50 based PyramidBox and vgg16 based SFD, which are denoted by {\em Pr50} and {\em Sv16} respectively. {All of these face detectors are trained on complete WIDER training set.}

\subsubsection{Evaluation metric}
Performance of face detection is commonly evaluated using average precision (AP) \cite{yang2016wider}, which computes the average precision value for recall value over 0 to 1. Higher values of AP correspond to better accuracy. However, AP is not a proper metric for evaluating the performance of adversarial perturbation. This is because AP penalizes mis-detections but is insensitive to false detections, therefore, it does not consider their effect on the obtained face set to be used to train AI based face synthesis methods. 

Because of this, we define a new metric, {\em data utility quality} (DUQ), to evaluate the utility of the obtained face set when used as training data for AI based face synthesis algorithms. We sum up the number of true, false detections and ground truth over all images as (\# True detections), (\# False detections) and (\# Ground truth faces) respectively. Specifically, DUQ is defined as
\begin{equation}
DUQ = \frac{\textrm{\# True detections} - \textrm{\# False detections}}{\textrm{\# Ground truth faces}}.   
\end{equation}
As the definition shows, DUQ takes value in the range  $(-\infty,1]$, where a value $1$ indicates all the faces are completely detected while no false detections are generated. On the other hand, a negative DUQ suggests a significant number of false positives have been included in the face set. A lower DUQ corresponds to a lower purity of the detection face set when used as training data for AI face synthesis algorithms, as such, a successful adversarial perturbation scheme should lead to smaller DUQ values.

We also use SSIM \cite{wang2004image} to assess the image visual quality after adversarial perturbation. SSIM takes value in $[0, 1]$, and higher value corresponds to better visual quality. 

\subsubsection{Implementation details and running time}
The proposed adversarial perturbation generation method is implemented with Python and package {\tt pyTorch}\footnote{Code will be made public upon acceptance of the paper.}. Some of the key constants are set as follows. The upper bound of distortion {(Mean Square Error)} is set to $\epsilon = 5 \times 10^{-5}$. The step size in each iteration is set to $\gamma_t =  \frac{30}{||\nabla L(\I_t)||_2}$.  The total number of iterations is $T = 200$. All experiments are performed on a machine which is equipped an Intel(R) Xeon(R) CPU E5-2620 v3 @ 2.40GHz with $24$ cores and $96$ GB RAM. The GPU we use is a NVIDIA TITAN X (Pascal) with $12$ GB memory. The average time for generating successful adversarial perturbation for an image is $4.69$ seconds.

\subsubsection{Baselines and compared methods} 
We evaluate the performance of our method and compared with three algorithms. The first is a simple baseline algorithm that adds random Gaussian noise to image and is denoted as {\em Random}. The second  algorithm is based on an adversarial perturbation generator for general object detectors \cite{xie2017adversarial}, and is denoted as {\em SSOD}. Specifically, SSOD attacks object detectors by reducing the confidence score of true detections. As SSOD was trained and evaluated on Pascal VOC dataset \cite{everingham2015pascal} which has no label corresponding to human faces, to facilitate comparison, we use the original code of SSOD and refine it on the face datasets in our experiments. The third algorithm we compare, denoted as {\em NNCO}, is the only existing white-box adversarial perturbation generation method dedicated to attack DNN based face detectors \cite{bose2018adversarial}. NNCO uses a GAN model with a generator of adversarial perturbations targeting vgg16 based Faster-RCNN face detectors trained and tested both on a same dataset 300-W. We compare our method with NNCO by both applying them on vgg16 face detector\footnote{There is no published code ready for use, so we re-implement their method following the descriptions in \cite{bose2018adversarial}.}.

\subsection{White-box Adversarial Perturbation Generation}

\begin{figure*}[t]
	\centering
	\includegraphics[width=0.9\linewidth]{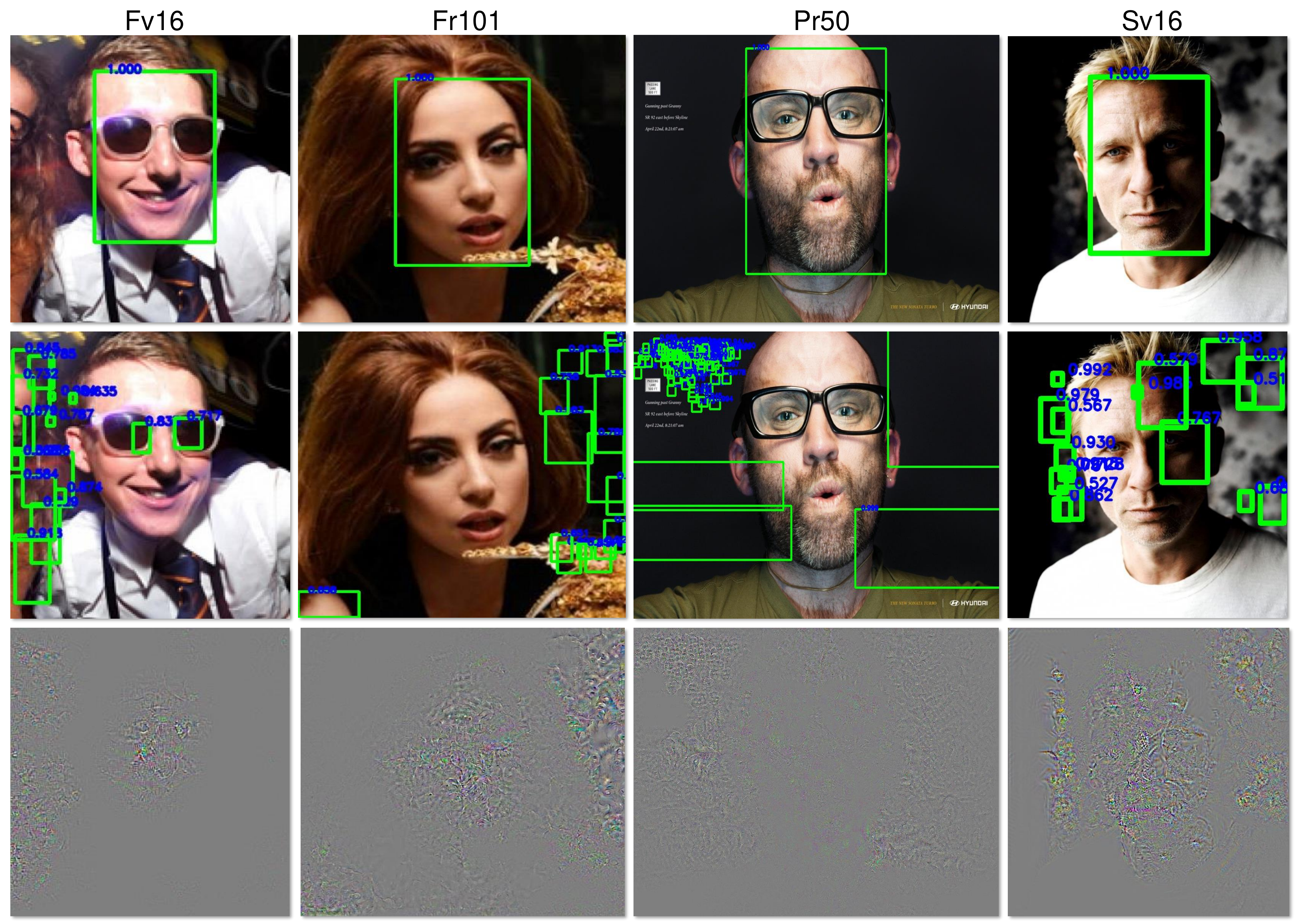}
	\vspace{-0.2cm}
	\caption{\em \small Visual examples of our method attacking Fv16, Fr101, Pr50 and Sv16 respectively. The top row corresponds to detection results on original images. The middle row corresponds to the detection results on images after adversarial perturbation are added to the original image. The bottom row show the actual noise added, which are amplified by 30 for better visualization. }
	\label{fig:demos}
	~\vspace{-1em}
\end{figure*}

Table \ref{table:fr} shows the performance of compared methods for Faster-RCNN and SSD based face detectors with different base networks before and after adversarial perturbations. We show both the effectiveness of the adversarial perturbation (DUQ) and image quality (SSIM). Figure \ref{fig:demos} provides four visual examples of the results of adversarial perturbation of face images. 

As these results show, face detection is significantly affected after the adversarial perturbations generated with our method are added to these images. For example, DUQ of Fv16 is reduced to $-4.89$ from $0.81$ on sub-WIDER dataset with a minor reduction of SSIM (0.02), this shows that our method conceal true detections in the original unperturbed image and introduce a large number of false detections (also, see Figure \ref{fig:demos}). The same trend is observed for AP scores, which drops to $4.2\%$ from $99.7\%$ using our method. In contrast, SSOD can only reduce DUQ of all face detectors on all datasets to around $0$, as it only considers reducing true detections. On the other hand, adding random Gaussian noise has almost no effect on face detectors, suggesting that the dependency structure in the adversarial perturbation is essential. The performance for the SSD based DNN face detectors are similar. 

\begin{figure}[t]
	\centering
	\includegraphics[width=0.95\linewidth]{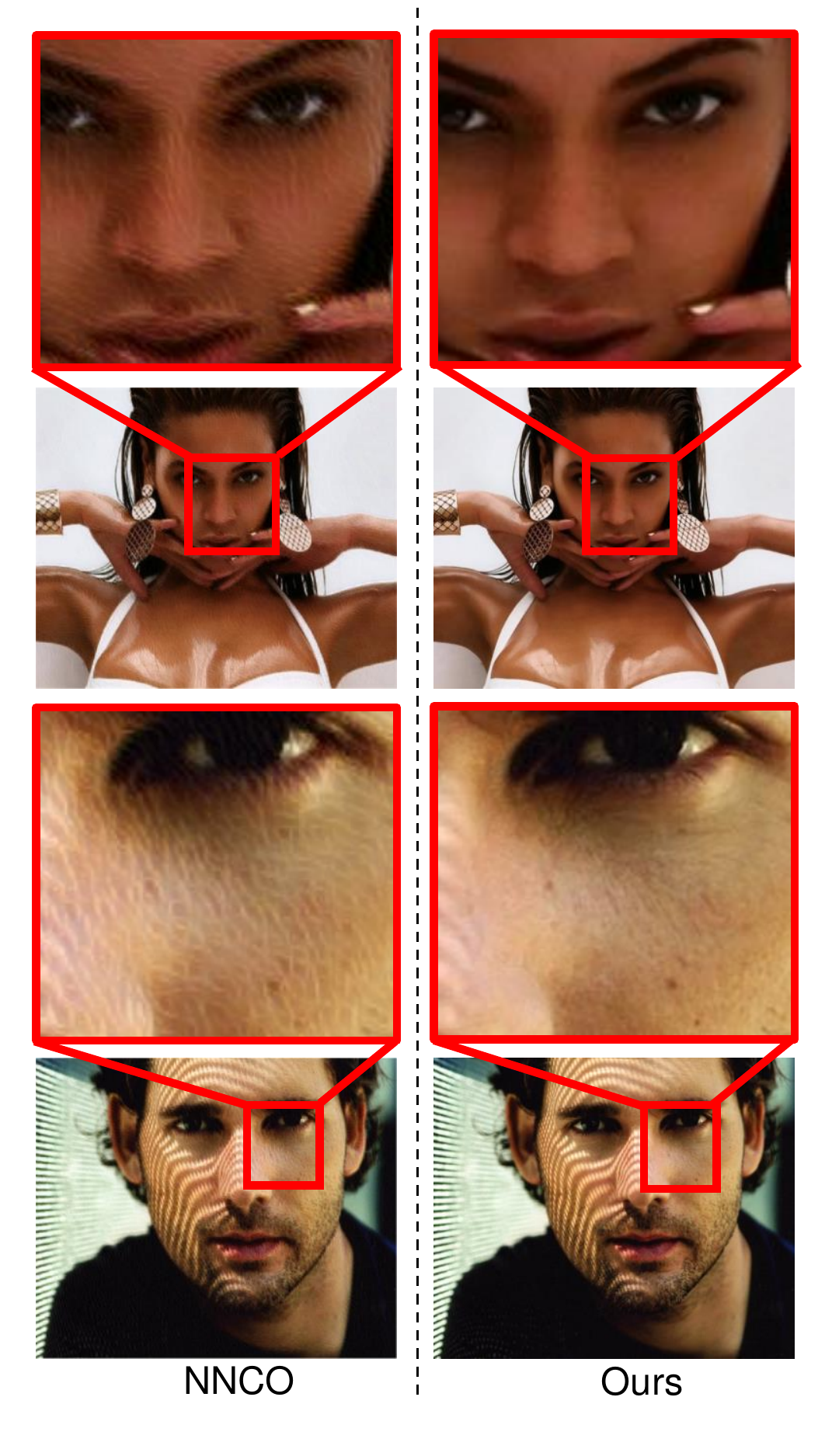}
	\vspace{-0.9cm}
	\caption{\em \small Visual comparison of the perturbed images between NNCO \cite{bose2018adversarial} and our method. We can the see the perturbed images generated by \cite{bose2018adversarial} has clear artifacts in the skin of faces, while the perturbations generated by our method are hardly to be perceived.}
	\label{fig:vis_compare}
	~\vspace{-1em}
\end{figure}

Furthermore, when compared with method NNCO, our method can reduce the DUQ of Fv16 from $0.81, 0.64, 0.55$ to $-4.89, -9.07, -7.52$ on three datasets respectively, while NNCO only reduces DUQ to $0.17, 0.06, 0.12$.  Moreover, compared with NNCO, our method achieves better image quality.  Figure \ref{fig:vis_compare} shows a visual comparison of the perturbed image generated by our method (top) and NNCO (bottom) with enlarged area. Note the visible artifacts generated by NNCO, which are not present in the perturbed image generated with our method. This is corroborated by the quantitative results when comparing the SSIM scores of our method ($0.98,0.98,0.98$) and those of NNCO ($0.92,0.91,0.92$) on all datasets, respectively 

However, adversarial perturbations generated under the white-box setting may not be able to extend to DNN based face detectors that use different base networks and different network parameters. This is confirmed by a set of experiments in which we use adversarial perturbation generated from one face detector to other different face detectors.  The performance is evaluated in DUQ, which is shown in Table \ref{table:single-transfer}. {{\em Ours$_{(\text{Fv16})}$}, {\em Ours$_{(\text{Fr101})}$}, {\em Ours$_{(\text{Pr50})}$} and {\em Ours$_{(\text{Sv16})}$} denote perturbed images generated by our method on Fv16, Fr101, Pr50 and Sv16 face detectors respectively.} The results show that the adversarial perturbation developed for one type of face detector can barely extend to other face detectors. 
		
\begin{table*}[t]
	\small
	\centering
	\caption{Performance of adversarial perturbation generation against Faster-RCNN and SSD based face detectors on three datasets. Fv16 and Fr101 denote Faster-RCNN based face detector with base network vgg16 \cite{simonyan2014very} and ResNet101 \cite{he2016deep}. Pr50 denotes PyramidBox \cite{tang2018pyramidbox} with base network ResNet50 and Sv16 denotes SFD \cite{zhang2017s3fd} with base network vgg16.}
	\begin{tabular}{|l|l|c|c|c|c|c|c|c|c|c|c|c|c|}
	    \hline
	    \multicolumn{2}{|c}{Adversarial Perturbation} & \multicolumn{4}{|c}{Sub-WIDER} & \multicolumn{4}{|c}{300-W} & \multicolumn{4}{|c|}{Sub-UMDFaces} \\
	    \cline{3-14}
		\multicolumn{2}{|c|}{Generation Method}    & Fv16    & Fr101     & Pr50    & Sv16   & Fv16    & Fr101     & Pr50    & Sv16    & Fv16    & Fr101     & Pr50    & Sv16   \\
		\hline 
		\multirow{4}{*}{DUQ} & Original        & 0.81       & 0.82     & 0.92      & 0.88      & 0.64       & 0.64      & 0.76        & 0.81    & 0.55       & 0.54      & 0.64        & 0.64 \\
		\cline{2-14}
		                     & Random          & 0.81       & 0.82     & 0.92      & 0.81      & 0.64       & 0.64      & 0.76        & 0.81    & 0.54       & 0.54      & 0.64        & 0.64 \\
		\cline{2-14}
		                     & NNCO \cite{bose2018adversarial} & 0.17       & -     & -      & -    & 0.06       & -  & - & - & 0.12 & - & - & - \\
		\cline{2-14}
		                     & SSOD \cite{xie2017adversarial}        & -0.06      & -0.21    & -0.74     & -0.53      & -0.02     & -0.03     & -0.81       & -1.01   & -0.18      & -0.65     & -0.88       & -0.99 \\
		\cline{2-14}
		                     & Ours            & \bf -4.89   & \bf -7.98 & \bf -19.18     & \bf -9.40  & \bf -9.07  & \bf -7.95  & \bf -17.91   & \bf -8.80 & \bf -7.52    & \bf -9.91  & \bf -26.88   & \bf -8.80 \\		                     
		\hline
	    \hline
		\multirow{2}{*}{SSIM} & NNCO \cite{bose2018adversarial} & 0.92       & -     & -      & -    & 0.91       & -  & - & - & 0.92 & - & - & - \\
		\cline{2-14}
		& SSOD \cite{xie2017adversarial}       & 1.0      & 0.99      & 0.97   & 0.99    & 0.94    & 0.98     & 0.95    & 0.98    & 1.0    & 1.0    & 0.95     & 0.98  \\
		\cline{2-14}
		                      & Ours            & 0.98     & 0.96      & 0.96   & 0.98    & 0.98    & 0.97     & 0.94    & 0.96    & 0.98   & 0.97   & 0.93     & 0.96 \\
		\hline
	\end{tabular}
	\label{table:fr}
\end{table*}

\begin{table}[t]
	\centering
	\small
	\caption{Performance evaluated in DUQ for attacking face detectors using different adversarial perturbation. Fv16 and Fr101 denote Faster-RCNN based face detector with base network vgg16 \cite{simonyan2014very} and ResNet101 \cite{he2016deep}. Pr50 denotes PyramidBox \cite{tang2018pyramidbox} with base network ResNet50 and Sv16 denotes SFD \cite{zhang2017s3fd} with base network vgg16.}
	\begin{tabular}{|l|c|c|c|c|}
		\hline
                                    & Fv16            & Fr101            & Pr50           & Sv16 \\
        \hline
        Original                    & 0.81            & 0.82             & 0.92           & 0.88          \\
		\hline
		Ours$_{(\text{Fv16})}$      & \bf -4.98       & 0.41             & 0.71           & 0.42          \\
		\hline
		Ours$_{(\text{Fr101})}$     & 0.66            & \bf -7.98        & 0.89           & 0.84          \\
		\hline
		Ours$_{(\text{Pr50})}$      & 0.71            & 0.74             & \bf -19.18     & 0.78          \\
		\hline
		Ours$_{(\text{Sv16})}$      & 0.02            & 0.58             & 0.32           & \bf -9.40     \\
		\hline		
	\end{tabular}
	\label{table:single-transfer}
\end{table}

\subsection{Gray-box Adversarial Perturbation Generation}
\label{sec:exp-gray-box}

\begin{figure}
    \centering
    \includegraphics[width=\linewidth]{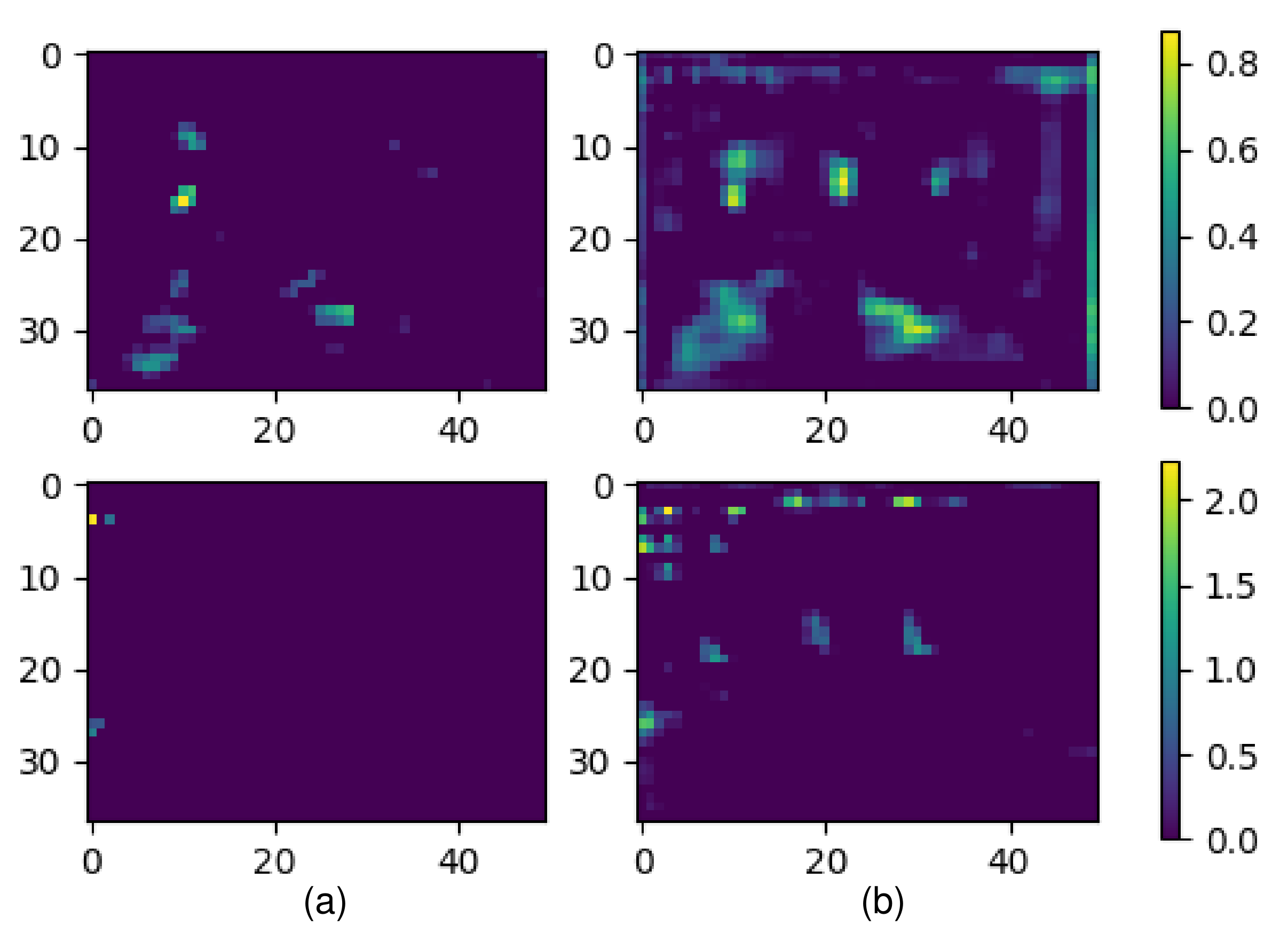}
    \vspace{-0.8cm}
    \caption{\em \small Illustration of feature map difference between (a) Fv16 and (b) Fv16$^*$ for the same input image. We select two channels (top and bottom row) from the output of base network for comparison.}
    \label{fig:feat-diff}
\end{figure}

\begin{table}[t]
	\centering
	\small
	\caption{Performance of gray-box attack for two refined DNN based face detectors. Fv16$^*$ and Fr101$^*$ denote Faster-RCNN based face detector with base network vgg16 \cite{simonyan2014very} and ResNet101 \cite{he2016deep}, which are trained on the union of WIDER \cite{yang2016wider} and FDDB \cite{fddbTech}. }
	\begin{tabular}{|l||c||c|c|c|}
		\hline
                              & SSIM  & Fv16$^*$      & Fr101$^*$ \\
		\hline
		Original              & 1.0   & 0.84              & 0.91              \\
		\hline
		Ours$^*_{(\text{Fv16})}$  & 0.98  & \bf -2.11         & 0.72              \\
		\hline
		Ours$^*_{(\text{Fr101})}$  & 0.96  & 0.78              & \bf -1.85         \\
		\hline	
		\end{tabular}
	\label{table:gray-box}
\end{table}
\begin{table*}[t]
	\centering
	\small
	\caption{Performance evaluated in DUQ for black-box adversarial perturbation generation. Rows denote perturbed images generated from different cases. Columns denote different face detectors. Fv16 and Fr101 denote Faster-RCNN based face detector with base network vgg16 \cite{simonyan2014very} and ResNet101 \cite{he2016deep}. Pr50 denotes PyramidBox face detector \cite{tang2018pyramidbox} with base network ResNet50 and Sv16 denotes SFD face detector \cite{zhang2017s3fd} with base network vgg16. The last three column Fr50, SSHv16, SSHr50 are black-box face detectors, which are Faster-RCNN based face detector with base network ResNet50 and SSH \cite{najibi2017ssh} with base network vgg16 and ResNet50.}
	\begin{tabular}{|l||c||c|c|c|c||c|c|c|}
		\hline
        \multirow{2}{*}{} & \multirow{2}{*}{SSIM}   & \multicolumn{4}{c||}{ known face detectors} & \multicolumn{3}{c|}{unknown face detectors} \\
        \cline{3-9}
                          &                         & Fv16    & Fr101    & Pr50   & Sv16  & Fr50   & SSHv16   & SSHr50 \\
        \hline
        Original   & 1.0  & 0.81      & 0.82      & 0.92       & 0.88       & 0.87    & 0.93    & 0.94 \\
        \hline
		SSOD$_{(\text{Fv16+Fr101+Pr50+Sv16})}$ & 0.95    & 0.14         & 0.34         & 0.26          & 0.04          & 0.57      & 0.65       & 0.66  \\
		\hline
        Ours$_{(\text{Fv16+Fr101+Pr50+Sv16})}$ & 0.91 & -4.94     & -5.69     & -14.14     & -6.37      & \bf -1.72   & \bf -0.26  & \bf -0.47 \\
		
		\hline	
	\end{tabular}
	\label{table:transfer}
\end{table*}

\begin{figure*}[t]
	\centering
	\includegraphics[width=0.8\linewidth]{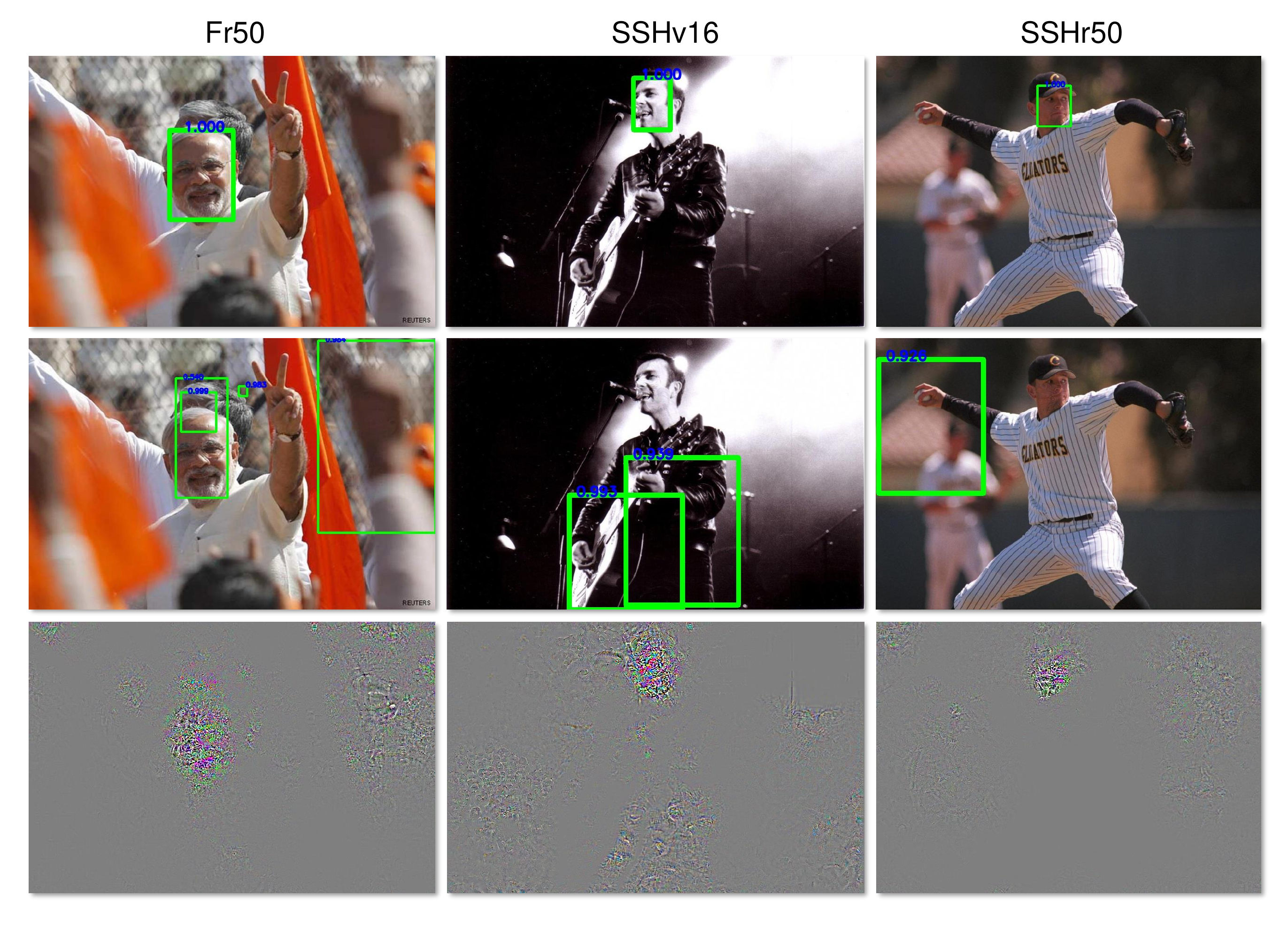}
	\vspace{-0.4cm}
	\caption{\em \small Visual examples of black-box attack on Fr50, SSHv16 and SSHr50 face detectors respectively. The top row corresponds to detection results on original images. The middle row corresponds to the detection results on images after adversarial perturbation are added to the original image. The bottom row show the actual noise added, which are amplified by 30 for better visualization. }
	\label{fig:black-box}
\end{figure*}

We next study adversarial perturbation generation under the gray-box setting, under which we assume knowing the network structure but not the parameters (weights) of the underlying DNN. This reflects the practical scenario where a pre-trained DNN based face detector is fine-tuned on different training dataset. To simulate this effect, we compare two DNN based face detectors, namely Faster-RCNN based face detector with base network vgg16 (Fv16) and ResNet101 (Fr101) trained on two different datasets, the first one is the original WIDER dataset, and the second one is the union of WIDER and FDDB \cite{fddbTech}\footnote{FDDB is another large face detection benchmark containing $5171$ faces in a set of $2845$ images.}. We denote the detectors trained on the augmented dataset as Fv16$^*$ and Fr101$^*$. Figure \ref{fig:feat-diff} shows feature map difference between (a) Fv16 and (b) Fv16$^*$ for the same input image. The feature map is extracted from the last convolution layer of the base network of Fv16 and Fv16$^*$ face detectors, which contains $512$ channels and we select two channels for comparison. Note the difference between the feature maps, reflecting that the parameters for the two base networks are not the same. 

We obtain the new perturbed images using the method developed in Section \ref{sec:method} according to Eq.\eqref{e:2a} for Fv16 and Fr101, which are denoted by {\em Ours$^*_{(\text{Fv16})}$}, {\em Ours$^*_{(\text{Fr101})}$} respectively. Table \ref{table:gray-box} shows the DUQ score on the two face detectors with different parameters, and indicates that the stochastic variant of our method can create perturbations that are robust to variations in parameters of the base network of the DNN based face detector. For example, the DUQ of Fv16$^*$ is reduced to $-2.11$ from $0.84$ using the adversarial perturbation by Ours$^*_{(\text{Fv16})}$, with slight change of image quality.

\subsection{Black-box Adversarial Perturbation Generation}
\label{sec:exp-black-box}

Following the method described in Section \ref{sec:black-box}, we generate adversarial perturbations using Eq.\eqref{e:4} for the combination of four face detectors for black-box attack of unknown face detectors. Specifically, we use the following three DNN based face detectors as the unknown face detectors:
\begin{itemize}
    \item Faster-RCNN based face detector with base network ResNet50 (denoted as {\em Fr50});
    \item SSD based face detector SSH \cite{najibi2017ssh} with base networks vgg16 (denoted as {\em SSHv16});
    \item SSD based face detector SSH \cite{najibi2017ssh} with base networks ResNet50 ({\em SSHr50}).
\end{itemize}
All the experiments are conducted on sub-WIDER dataset. We denote our adversarial perturbation generation method targeting the ensemble of known face detectors as {\em Ours$_{(\text{Fv16+Fr101+Pr50+Sv16})}$} as described in Section \ref{sec:black-box}. For comparison, we adapt the adversarial perturbation generation scheme for general object detectors \cite{xie2017adversarial} to face detectors, and denote the corresponding method as {\em SSOD$_{(\text{Fv16+Fr101+Pr50})}$}. Unlike our method, SSOD can only be extended to the black-box setting by generating adversarial perturbations for different face detectors independently, and then uses their summation as the final perturbations to any unknown face detectors. On the other hand, our method is based on an optimization problem, Eq. \eqref{e:4}, and also considers false detections. 

Table \ref{table:transfer} shows the performance evaluated in DUQ of SSOD and our method. In addition, Figure \ref{fig:black-box} shows three examples of black-box attack on Fr50, SSHv16 and SSHr50. As the adversarial perturbations are obtained by considering the four known DNN based face detectors, they are expected to work well in those cases for all methods, which is confirmed by their performance under the white-box setting.

When applied to unknown DNN based face detectors, both methods tend to be effective in reducing DUQ of the resulting face set. However, our method shows more reduction and usually leads to negative DUQ scores, suggesting that it generates more false detections, and thus is more effective in reducing the quality of the face set as training data for AI face synthesis system. This is further corroborated by the visual results shown in Figure \ref{fig:black-box}.

\subsection{Robustness}
\label{sec:robust}
We further evaluate the robustness of our adversarial perturbation generation method with regards to several types of image processing operations including {\em JPEG compression}, {\em additive noise} and {\em  blurring}. The results are obtained on the sub-WIDER dataset.

\subsubsection{JPEG Compression} 
We generate perturbed images then JPEG compressed them with different quality factors $[60:100]$. Figure \ref{fig:robust} (top) illustrates DUQ performance of face detectors on perturbed images with compression degree decreasing. The blue line denotes original images and the red line denotes perturbed images. We observe that the JPEG compression to original images does not affect DUQ performance, but the JPEG compression to perturbed images degrades the attacking performance. Compared to Fr101 and Pr50, the curves for Fv16 and Sv16 degrade more slowly, which indicates these two face detectors are affected by adversarial perturbation even after JPEG compression. It is probably because Fr101 and Pr50 are more robust to adversarial perturbation attack than Fv16 and Sv16. 

\subsubsection{Additive White Noise}
We also evaluate the robustness of our method with regards to additive white Gaussian noise ${\cal N}(0,\sigma_\text{add})$, where the standard deviation $\sigma_\text{add}$ is set to $[0:10]$. $\sigma_\text{add} = 0$ denotes no additive noise is added. The effectiveness of additive noise is shown in Figure \ref{fig:robust} (middle). We observe the DUQ performance of face detectors on perturbed images is recovered slowly with additive noise scale increased, which indicates the adversarial perturbations have ability to resist the disruption of additive noise. Similar to the trends in JPEG compression, Fv16 and Sv16 raises more smoothly compared to Fr101 and Pr50. 

\subsubsection{Blurring}
\begin{figure*}[t]
	\centering
	\includegraphics[width=0.93\linewidth]{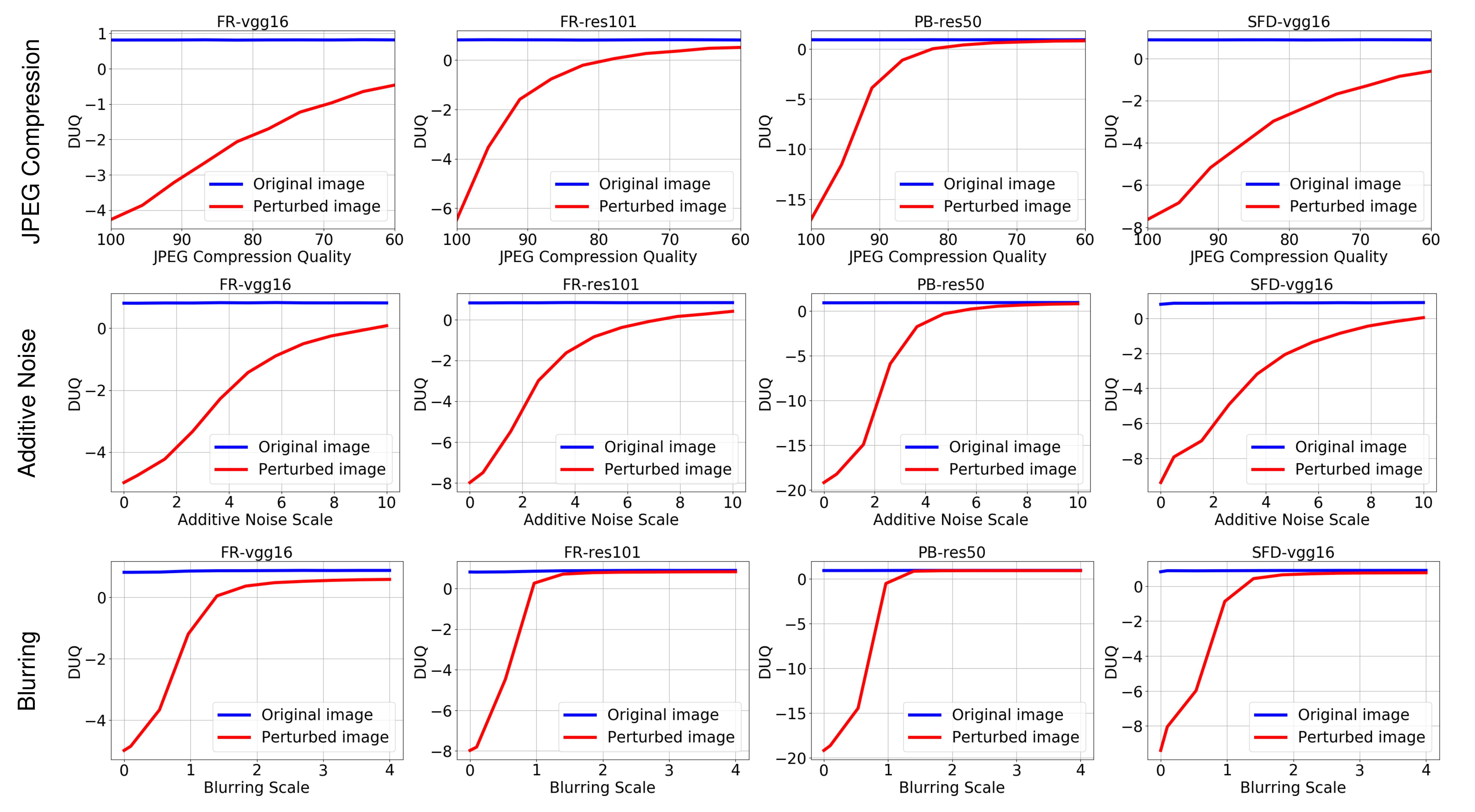}
	\vspace{-0.5cm}
	\caption{\em \small The effect of JPEG compression, additive noise and blurring on DUQ performance of each face detector. See text for more details.}
	\label{fig:robust}
\end{figure*}

Gaussian blurring smooths the image using Gaussian blurring kernel, which could affect the adversarial perturbation attack. In our case, we use kernel ${\cal N}(0,\sigma_\text{blur})$ with size $5 \times 5$ for blurring. For different blurring scales, we set the standard deviation $\sigma_\text{blur}$ to $[0:4]$, where $\sigma_\text{blur} = 0$ denotes no blurring is applied. The effectiveness of blurring is shown in Figure \ref{fig:robust} (bottom), which reveals the attacking ability of adversarial perturbation is degraded with $\sigma_\text{blur}$ increasing and has almost no effect when $\sigma_\text{blur} > 2$.

\section{Conclusion}
\label{sec:conclusion}
AI synthesized fake faces are becoming a problem encroaching our trust to online media. As most AI based face synthesis algorithms require automatic face detection as an indispensable pre-processing step in preparing training data, an effective protection scheme can be obtained by disrupting the face detection methods. In this work, we develop a {\em proactive} protection method to deter bulk reuse of automatically detected face for the production of AI synthesized faces. Our method exploits the sensitivity of DNN based face detectors and use adversarial perturbation to contaminate the face sets.  This is achieved by generating imperceptible adversarial perturbations to disrupt face detectors. We describe attacking schemes for white-box, gray-box and black-box settings, and empirically show the effectiveness of our methods in disrupting state-of-the-art DNN based face detectors on several datasets.

We expect this technology to solicit counter-measures from the forgery makers. In particular, operations that can destroy or reduce the adversarial perturbation are expected to be developed. It is thus our continuing effort to improve the robustness of the adversarial perturbation generation method. Another important direction to further explore is a more generic black-box attack scheme that does not limit to DNN based face detectors and do not rely on the differentialbility of the underlying model. Furthermore, we will also work on improving the running time efficiency of the current method so it can scale up to large number of images. 

\bibliographystyle{IEEEtran}
\bibliography{ref}
\end{document}